\documentclass[graybox]{svmult}
\usepackage[utf8]{inputenc}                
\usepackage{mathrsfs}                                              
\usepackage[T1]{fontenc}
\usepackage{listings}
\usepackage{color}
\usepackage{bm}
\usepackage{url}
\usepackage{cite}
\usepackage{epsfig}
\usepackage{mathtools}
\usepackage{subfigure}
\usepackage{makecell}
\usepackage{multirow}
\usepackage{footnote}
\DeclareGraphicsExtensions{.pdf,.png,.jpg,.fig}

\usepackage[cmintegrals]{newtxmath}
\usepackage[width=125mm,left=10mm,paperwidth=146mm,height=193mm,top=11mm,paperheight=217mm, lmargin=19mm, rmargin=19mm, bmargin=15mm]{geometry}
\usepackage[colorlinks=true,citecolor=red,linkcolor=blue]{hyperref}

\makeatletter
\newcommand\footnoteref[1]{\protected@xdef\@thefnmark{\ref{#1}}\@footnotemark}
\makeatother

\newcommand{\eg}{\textit{e.g.}}
\newcommand{\ie}{\textit{i.e.}}

\newcommand{\alexnet}{{AlexNet }}
\newcommand{\vgg}{VGG}
\newcommand{\resnet}{{ResNet}}

\newcommand{\googlenet}{{GoogLeNet }}
\newcommand{\shufflenet}{{ShuffleNet}}
\newcommand{\mobilenet}{{MobileNet}}
\newcommand{\mnasnet}{{MNASNet}}
\newcommand{\inception}{{Inception}}

\makeindex 
\begin{document}

\title*{Computer-Aided Road Inspection: Systems and Algorithms}
\author{Rui Fan, Sicen Guo, Li Wang, Mohammud Junaid Bocus}
\institute{Rui Fan \at Tongji University, \email{rfan@tongji.edu.cn}
\and Sicen Guo \at Tongji University, \email{guosicen@tongji.edu.cn}
\and Li Wang \at Continental AG, \email{li.wang@ieee.org}
\and Mohammud Junaid Bocus \at University of Bristol, \email{junaid.bocus@bristol.ac.uk}
}

\maketitle

\abstract{ 
Road damage is an inconvenience and a safety hazard, severely affecting vehicle condition, driving comfort, and traffic safety. The traditional manual visual road inspection process is pricey, dangerous, exhausting, and cumbersome. Also, manual road inspection results are qualitative and subjective, as they depend entirely on the inspector's personal experience. Therefore, there is an ever-increasing need for automated road inspection systems. This chapter first compares the five most common road damage types. Then, 2-D/3-D road imaging systems are discussed. Finally, state-of-the-art machine vision and intelligence-based road damage detection algorithms are introduced. 
}

\clearpage

\section{Introduction}
\label{sec.introduction}

The condition assessment of concrete and asphalt civil infrastructures (\eg, tunnels, bridges, and pavements) is essential to ensure their serviceability while still providing maximum safety for the users \cite{fan2018road}. It also allows the government to allocate limited resources for infrastructure maintenance and appraise long-term investment schemes \cite{koch2015review}. The detection and reparation of road damages (\eg, potholes, and cracks) is a crucial civil infrastructure maintenance task because they are not only an inconvenience but also a safety hazard, severely affecting vehicle condition, driving comfort, and traffic safety \cite{fan2019pothole}. 

Traditionally, road damages are regularly inspected (\ie, detected and localized) by certified inspectors or structural engineers \cite{kim2014review}. However, this manual visual inspection process is time-consuming, costly, exhausting, and dangerous \cite{fan2020we}. Moreover, manual visual inspection results are qualitative and subjective, as they depend entirely on the individual's personal experience \cite{mathavan2015review}. Therefore, there is an ever-increasing need for automated road inspection systems, developed based on cutting-edge machine vision and intelligence techniques \cite{fan2021rethinking}. 

A computer-aided road inspection system typically consists of two major procedures \cite{fan2021eusipco}: (1) road data acquisition, and (2) road damage detection (\ie, recognition/segmentation and localization). The former typically employs active or passive sensors (\eg, laser scanners \cite{tsai2018pothole}, infrared cameras \cite{jahanshahi2013unsupervised}, and digital cameras \cite{fan2019cvprw}) to acquire road texture and/or spatial geometry, while the latter commonly uses 2-D image analysis/understanding algorithms, such as image classification \cite{zhang2016road}, semantic segmentation \cite{fan2021graph}, object detection \cite{dhiman2019pothole}, and/or 3-D road surface modeling algorithms \cite{fan2019pothole} to detect the damaged road areas. 

This chapter first compares the most common types of road damages, including crack, spalling, pothole, rutting, and shoving. Then, various technologies employed to acquire 2-D/3-D road data are discussed. Finally, state-of-the-art (SOTA) machine vision and intelligence approaches, including 2-D image analysis/understanding algorithms and 3-D road surface modeling algorithms, developed to detect road damages are introduced.

\section{Road Damage Types}
\label{sec.road_damage_type}
\begin{figure}[!h]
	\begin{center}
		\centering
		\includegraphics[width=0.70\textwidth]{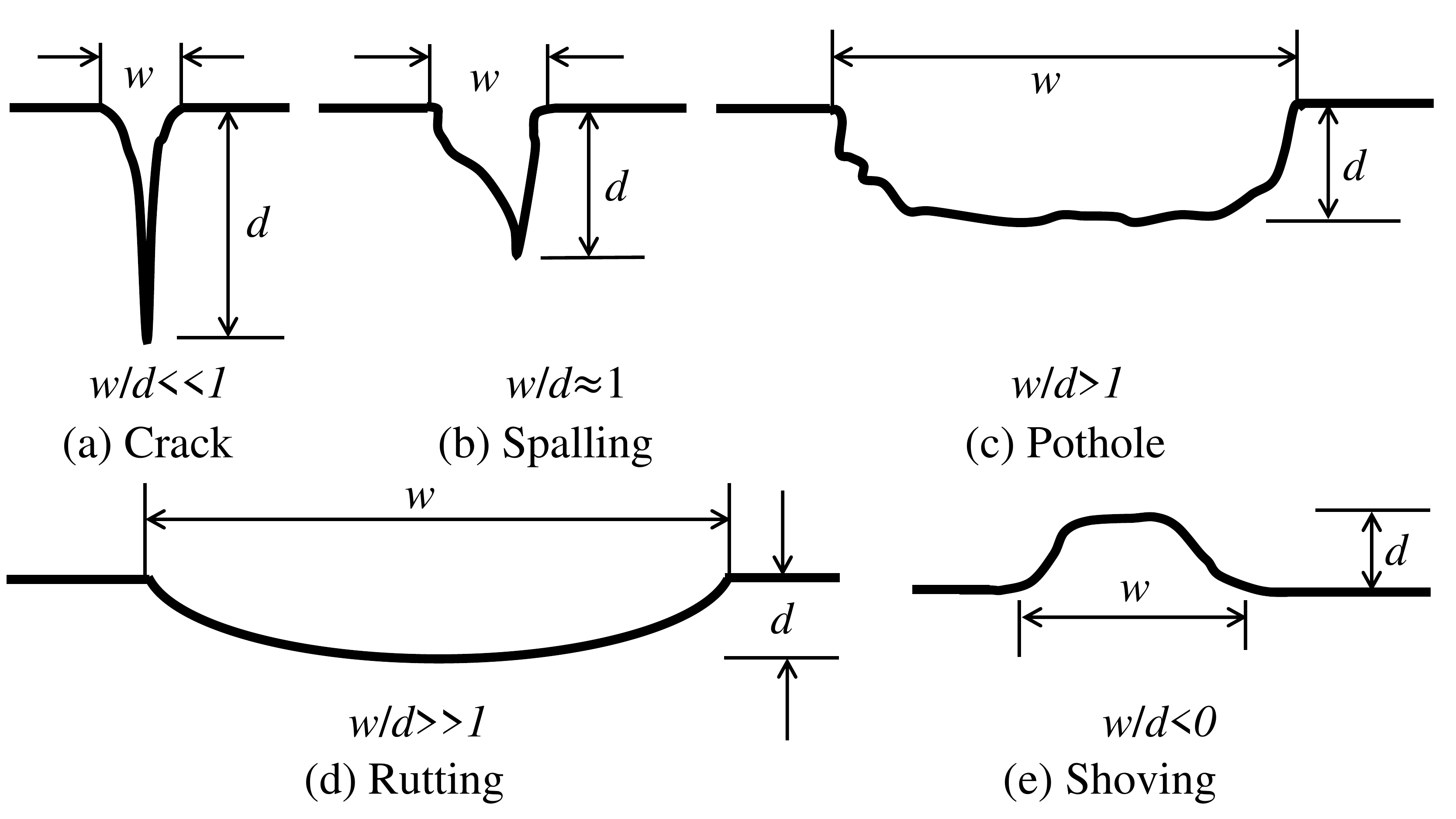}
		\centering
		\caption{Five common types of road damages, where $w$ and $d$ represent the lateral and depth magnitudes, respectively. }
		\label{fig.road_damage_type}
	\end{center}
\end{figure}

Crack, spalling, pothole, rutting, and shoving are five common types of road damages \cite{mathavan2015review}. Their structures are illustrated in Fig. \ref{fig.road_damage_type}. 
Road crack has a much larger depth when compared to its dimensions on the road surface, presenting a unique challenge to the imaging systems; Road spalling has similar lateral and depth magnitudes, and thus, the imaging systems designed specifically to measure this type of road damages should perform similarly in both lateral and depth directions; Road pothole is a considerably large structural road surface failure, measurable by an imaging setup with a large field of view; Road rutting is extremely shallow along its depth, requiring a measurement system with high accuracy in the depth direction; Road shoving refers to a small bump on the road surface, which makes its profiling with some imaging systems difficult.

\section{Road Data Acquisition}
\label{sec.road_data_collection}
\subsection{Sensors}

The use of 2-D imaging technologies (\ie, digital imaging or digital image acquisition) for road data collection started as early as 1991 \cite{mahler1991pavement, koutsopoulos1993primitive}. However, the spatial structure cannot always be explicitly illustrated in 2-D road images \cite{fan2018road}. Moreover, the image segmentation algorithms performing on either gray-scale or color road images can be severely affected by various factors, most notably by poor illumination conditions \cite{jahanshahi2013unsupervised}. Therefore, many researchers \cite{fan2018road, fan2018real, fan2019cvprw, fan2021rethinking} have resorted to 3-D imaging technologies, which are more feasible to overcome the disadvantages mentioned above and simultaneously provide an enhancement of road damage detection accuracy. 
 
The first reported effort \cite{laurent1997road} on leveraging 3-D imaging technology for road data collection can be dated back to 1997. This section discusses the existing 3-D imaging technologies applied for road data acquisition.

\begin{figure}[!h]
	\begin{center}
		\centering
		\includegraphics[width=0.75\textwidth]{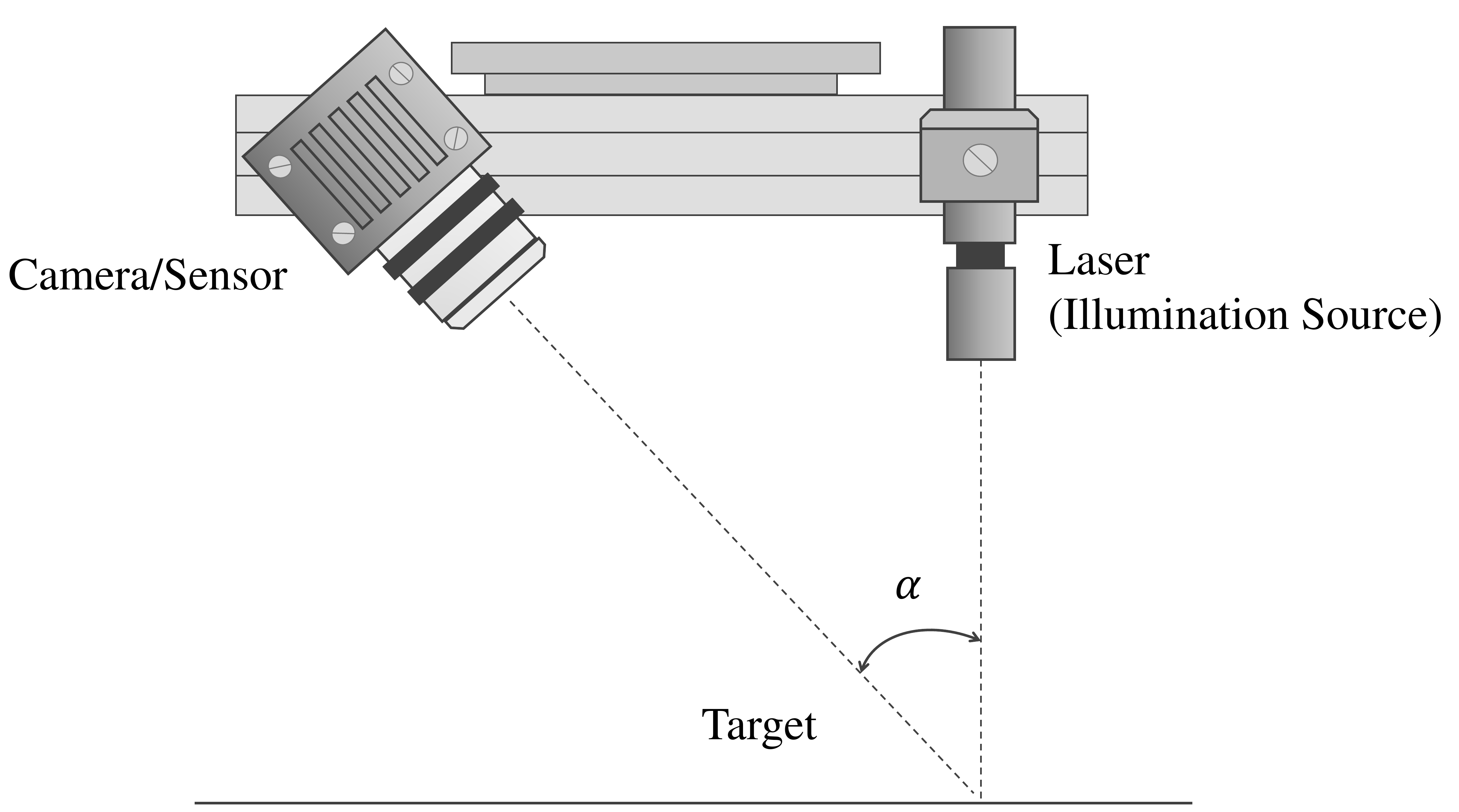}
		\centering
		\caption{Laser triangulation. The camera/sensor is located at a known distance from the laser’s illumination source. Therefore, accurate point measurements can be made by calculating the reflection angle of the laser light.}
		\label{fig.laser}
	\end{center}
\end{figure}

Laser scanning is a well-established imaging technology for accurate 3-D road geometry information acquisition \cite{mathavan2015review}. Laser scanners are developed based on trigonometric triangulation, as illustrated in Fig. \ref{fig.laser}. 
The sensor is located at a known distance from the laser’s source, and therefore, accurate point measurements can be made by calculating the reflection angle of the laser light. Auto-synchronized triangulation \cite{laurent1997road} is a popular variation of classic trigonometric triangulation and has been widely utilized in laser scanners to capture the 3-D geometry information of near-flat road surfaces \cite{mathavan2015review}. However, laser scanners have to be mounted on dedicated road inspection vehicles, such as a Georgia Institute of Technology Sensing Vehicle \cite{tsai2018pothole}, for  3-D road data collection, and such vehicles are not widely used because they involve high-end equipment and their long-term maintenance can be costly as well \cite{fan2021rethinking}. 

\begin{figure}[!h]
	\begin{center}
		\centering
		\includegraphics[width=0.70\textwidth]{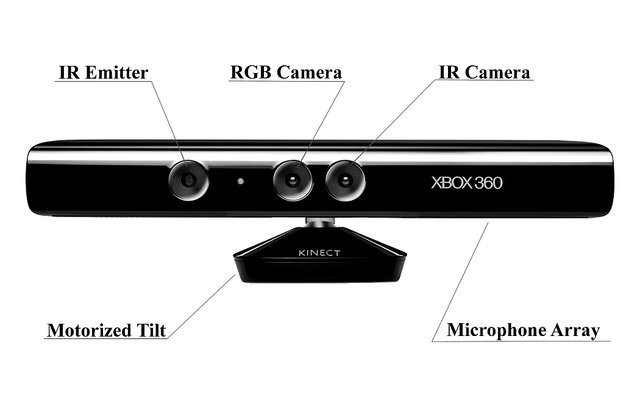}
		\centering
		\caption{Microsoft Kinect sensor \cite{barbatoreal}. }
		\label{fig.kinect}
	\end{center}
\end{figure}

Microsoft Kinect sensors \cite{jahanshahi2013unsupervised}, as illustrated in Fig. \ref{fig.kinect}, were initially designed for the Xbox-360 motion sensing games. Microsoft Kinect sensors are typically equipped with an RGB camera, an infrared (IR) sensor/camera, an IR emitter, microphones, accelerometers, and a tilt motor for motion tracking facility \cite{mathavan2015review}. The working range of Microsoft Kinect sensors is 800-4000 mm, making it suitable for road imaging when mounted on a vehicle. There are three reported efforts on 3-D road data acquisition and damage detection using Microsoft Kinect sensors \cite{joubert2011pothole, jahanshahi2013unsupervised, moazzam2013metrology}. However, it is also reported that Microsoft Kinect sensors greatly suffer from IR saturation in direct sunlight \cite{fan2019pothole}.

\begin{figure}[!h]
	\begin{center}
		\centering
		\includegraphics[width=0.55\textwidth]{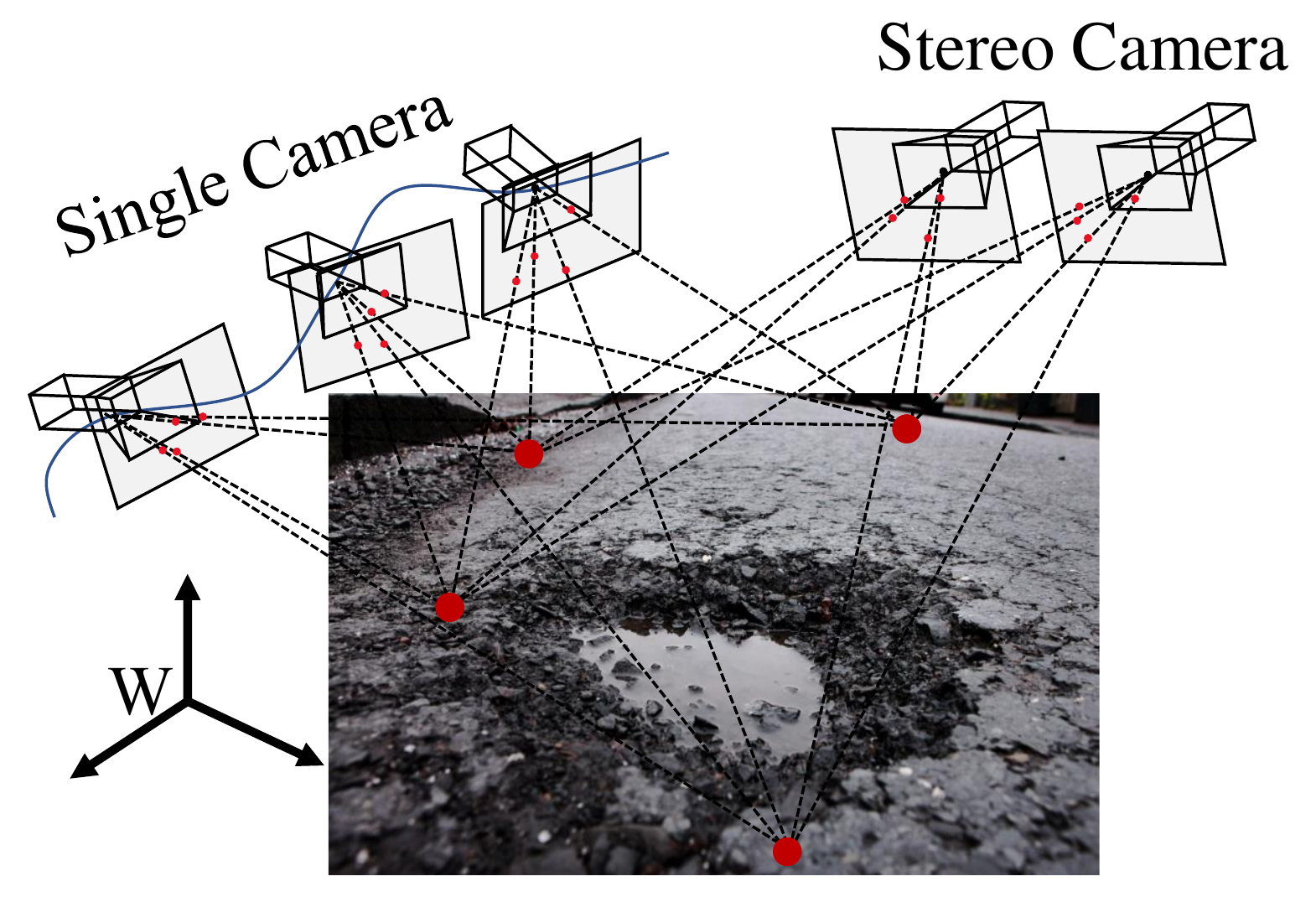}
		\centering
		\caption{3-D road imaging with camera(s) \cite{ma2022computer}. 
		}
		\label{fig.epiploar}
	\end{center}
\end{figure}

The 3-D geometry of a road surface can also be reconstructed using multiple images captured from different views \cite{fan2018road}. The theory behind this 3-D imaging technique is typically known as \textit{multi-view geometry} \cite{andrew2001multiple}. These images can be captured using a single movable camera \cite{jog2012pothole} or an array of synchronized cameras \cite{fan2019cvprw}. In both cases, dense correspondence matching (DCM) between two consecutive road video frames or between two synchronized stereo road images is the key problem to be solved, as illustrated in Fig. \ref{fig.epiploar}. Structure from motion (SfM) \cite{ullman1979interpretation} and optical flow (OF) estimation \cite{wang2020corl} are the two most commonly used techniques for monocular DCM, while stereo vision (also known as binocular vision, stereo matching, or disparity estimation) \cite{fan2018road, fan2021rethinking} is typically employed for binocular DCM.  SfM methods estimate both camera poses and the 3-D points of interest from road images captured from multiple views, linked by a collection of visual features. They also leverage bundle adjustment
(BA) \cite{triggs1999bundle} algorithm to refine the estimated camera poses and 3-D point locations by minimizing a cost function known as total re-projection error \cite{bhutta2020loop}. 
OF describes the motion of pixels between consecutive frames of a video sequence \cite{wang2021end}. Stereo vision acquires depth information by finding the horizontal positional differences (disparities) of the visual feature correspondence pairs between two synchronously captured road images \cite{fan2018road}. The traditional stereo vision algorithms formulate disparity estimation as either a local block matching problem \cite{fan2018icia} or a global/semi-global energy minimization problem (solvable with various Markov random field-based optimization techniques) \cite{hirschmuller2007stereo, sun2003stereo}, while data-driven algorithms typically solve stereo matching with convolutional neural networks (CNNs) \cite{wang2021pvstereo}. Despite the low cost of digital cameras, DCM accuracy is always affected by various factors, most notably by poor illumination conditions \cite{fan2019pothole}. Furthermore, it is always tricky for monocular DCM algorithms to recover the absolute scale of a 3-D road geometry model without considering the complicated environmental hypotheses \cite{fan2021eusipco}. On the other hand, the feasibility of binocular DCM algorithms relies on the well-conducted stereo rig calibration \cite{fan2018road}, and therefore, some systems incorporated stereo rig self-calibration functionalities to ensure that their captured stereo road images are always well-rectified.  

In addition to the aforementioned three common types of 3-D imaging technologies, shape from focus (SFF) \cite{danzl2009focus, sun2005autofocusing, nayar1994shape}, shape from defocus (SFDF) \cite{wohler2013triangulation, kuhl2006monocular, subbarao1994depth}, shape from shading (SFS) \cite{nayar1990surface}, photometric stereo \cite{woodham1980photometric, barsky20034}, interferometry \cite{fujimoto2000optical, walecki2006determining}, structured light imaging \cite{scharstein2003high, muzet2009surface}, and time-of-flight (ToF) \cite{oggier2004all, anderson2005experimental} are other alternatives for 3-D geometry reconstruction. However, they were not widely used for 3-D road data acquisition. The interested reader is referred to \cite{mathavan2015review} for a detailed description of these technologies (both theoretical and practical aspects are covered).

\subsection{Public Datasets}
\label{sec.datasets}
Road damage detection datasets are mainly created for crack or pothole detection. This section details the commonly used crack and pothole detection datasets. 

\subsubsection{Crack Detection Datasets}
\label{sec.crack_dataset}
The Middle East Technical University (METU) dataset \cite{Oezgenel2018} was created using the method proposed in \cite{zhang2016road} for image classification. The original road images (resolution: 4032$\times$3024 pixels) were taken at the METU. It is divided into two classes: negative and positive. Each class has 20,000 cropped road images (resolution: 227$\times$227 pixels). This dataset is publicly available at \url{data.mendeley.com/datasets/5y9wdsg2zt/2}. 

SDNET2018 \cite{Maguire2018} is another large-scale road crack detection dataset containing over 56,000 concrete road images (resolution: 256$\times$256 pixels) of bridge decks, walls, and pavements, with crack length ranging from 0.06 mm to 25 mm. Because of its diversity in crack size and width, it is commonly used for network evaluation. This dataset can be accessed at \url{digitalcommons.usu.edu/all datasets/48}.

\subsubsection{Pothole Detection Datasets}
\label{sec.pothole_dataset}
Bristol pothole detection dataset \cite{zhang2013advanced} was collected using two PointGrey Flea 3 cameras, mounted in parallel onto the optical rail, with an Arduino synchronization board in the middle. This produces a 20Hz clock signal to synchronize the cameras. The lenses on the cameras are wide-angle lenses with 2.8 mm focal length and F/1.2 aperture. They have a 93.2$^\circ\times$70.7$^\circ$ field of view, with the forward point setup. A threshold of $0.04$ m was set to detect the pothole point cloud.

\cite{fan2019pothole} provides the world-first multi-modal road pothole detection dataset containing 67 groups of RGB images, disparity images, transformed disparity images,  pixel-level annotations, and 3-D point clouds. This dataset can be used to design and evaluate semantic segmentation networks. This dataset is publicly available at \url{github.com/ruirangerfan/stereo_pothole_datasets}. 

The Pothole-600 dataset \cite{fan2020we} was created using a ZED stereo camera. It contains 600 collections of RGB images, transformed disparity images, and pixel-level annotations. This dataset can be employed to train both single-modal and data-fusion semantic segmentation networks, as illustrated in Fig. \ref{fig.semantic_seg}. It is available at \url{sites.google.com/view/pothole-600/}.

Cranfield PotDataset \cite{Alzoubi2018} was created using a DFK 33UX264 colour camera, collecting road images (resolution: 1024$\times$768) at 30 fps. This dataset contains 1610 pothole images, 1664 manhole images, 2703 asphalt pavement images, 2365 road marking images, and 2533 shadow images. This dataset can be used to train and evaluate image classification networks. This dataset is publicly available at \url{cord.cranfield.ac.uk/articles/dataset/PotDataset/5999699/1}.

\section{Road Damage Detection}
\label{sec.road_damage_detection}
The SOTA road damage detection approaches are developed based on either 2-D image analysis/understanding or 3-D road surface modeling. The former methods typically utilize traditional image processing algorithms \cite{jahanshahi2013unsupervised, fan2019road} or modern CNNs \cite{fan2020we,dhiman2019pothole} to detect road damage, by performing either pixel-level image segmentation or instance-level object recognition on RGB or depth/disparity images. The latter approaches typically formulate the 3-D road point cloud as a planar/quadratic surface \cite{fan2019pothole, zhang2013advanced, ozgunalp2016vision}, whose coefficients can be obtained by performing robust surface modeling. The damaged road areas can be detected by comparing the difference between the actual and modeled road surfaces. 

\begin{figure}[!h]
	\begin{center}
		\centering
		\includegraphics[width=0.9\textwidth]{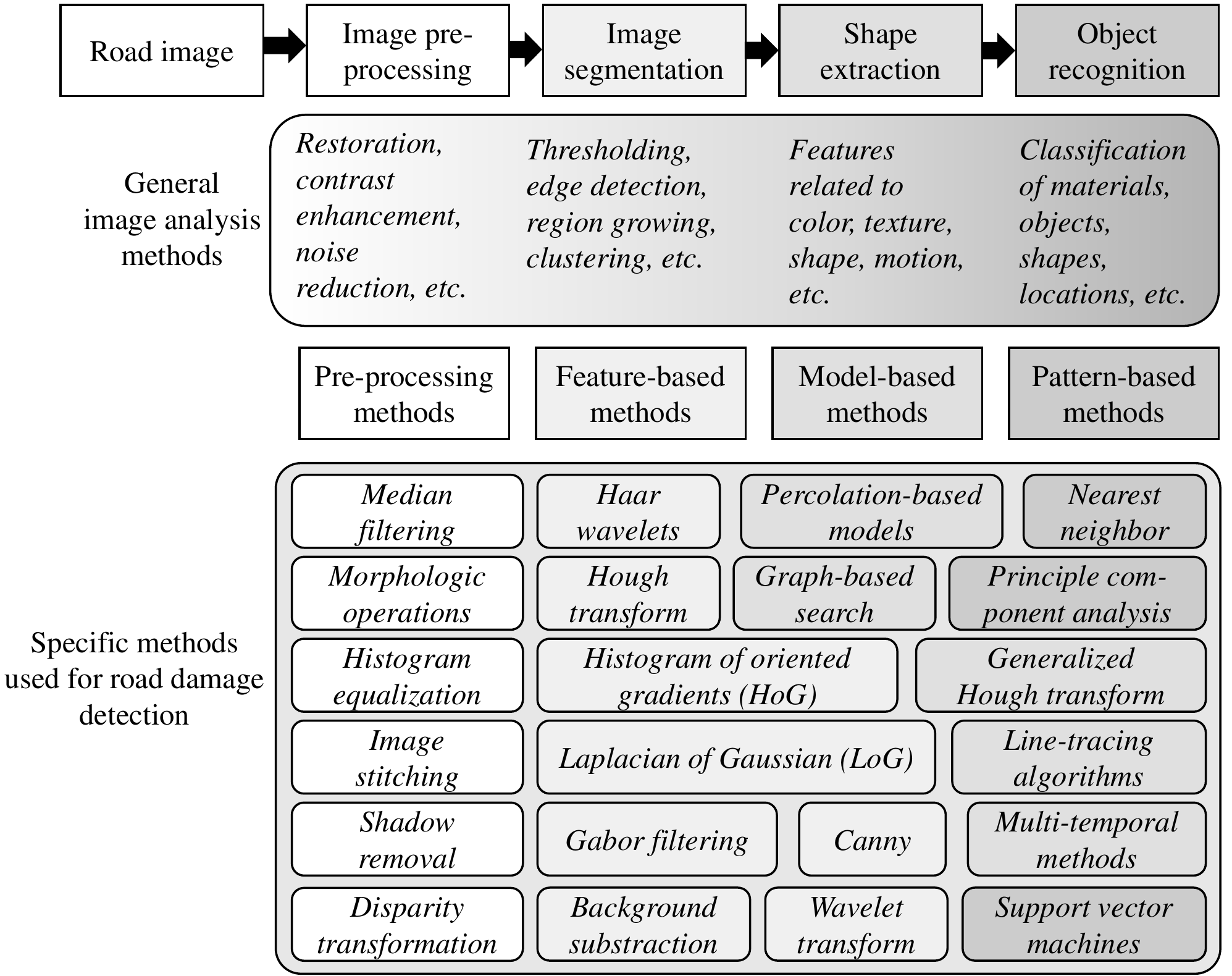}
		\centering
		\caption{An overview of the traditional image analysis-based road damage detection approaches. }
		\label{fig.vision_methods}
	\end{center}
\end{figure}

\subsection{2-D Image Analysis/Understanding-Based Approaches}

\subsubsection{Traditional image analysis-based approaches}
\label{sec.traditional_approaches}
An overview of the traditional image analysis-based road damage detection approaches is illustrated in Fig. \ref{fig.vision_methods}. These approaches typically consist of four main procedures: 1) image pre-processing, 2) image segmentation, 3) shape extraction, and 4) object recognition \cite{buza2013pothole}. \cite{ryu2015image} is a classic example of this algorithm type. It first uses a histogram-based thresholding algorithm \cite{koch2011pothole} to segment (binarize) the input gray-scale road images. The segmented road images are then processed with a median filter and a morphology operation to further reduce the redundant noise. Finally, the damaged road areas are extracted by analyzing a pixel intensity histogram. Additionally, \cite{koch2011pothole} solved road pothole detection with similar techniques. It first applies the triangle algorithm \cite{zack1977automatic} to segment gray-scale road images. An ellipse then models the boundary of an extracted potential road pothole area. Finally, the image texture within the ellipse is compared with the undamaged road area texture. If the former is coarser than the latter, the ellipse is considered to be a pothole.

However, both color and gray-scale image segmentation techniques are severely affected by various factors, most notably by poor illumination conditions \cite{fan2019pothole}. Therefore, some researchers performed segmentation on depth/disparity images, which has shown to achieve better performance in terms of separating damaged and undamaged road areas \cite{jahanshahi2013unsupervised,tsai2018pothole,fan2019road}. For instance, \cite{jahanshahi2013unsupervised} acquires the depth images of pavements using a Microsoft Kinect sensor. The depth images are then segmented using wavelet transform algorithm \cite{beylkin2009fast}. For effective operation, the Microsoft Kinect sensor has to be mounted perpendicularly to the pavements. In 2018, \cite{tsai2018pothole} proposed to detect road potholes from depth images acquired by a highly accurate laser scanner mounted on a Georgia Institute of Technology sensing vehicle. The depth images are first processed with a high-pass filter so that the depth values of the undamaged road pixels become similar. The processed depth image is then segmented using watershed method \cite{najman1994watershed} for road pothole detection. 

\begin{figure}[!h]
	\begin{center}
		\centering
		\includegraphics[width=0.95\textwidth]{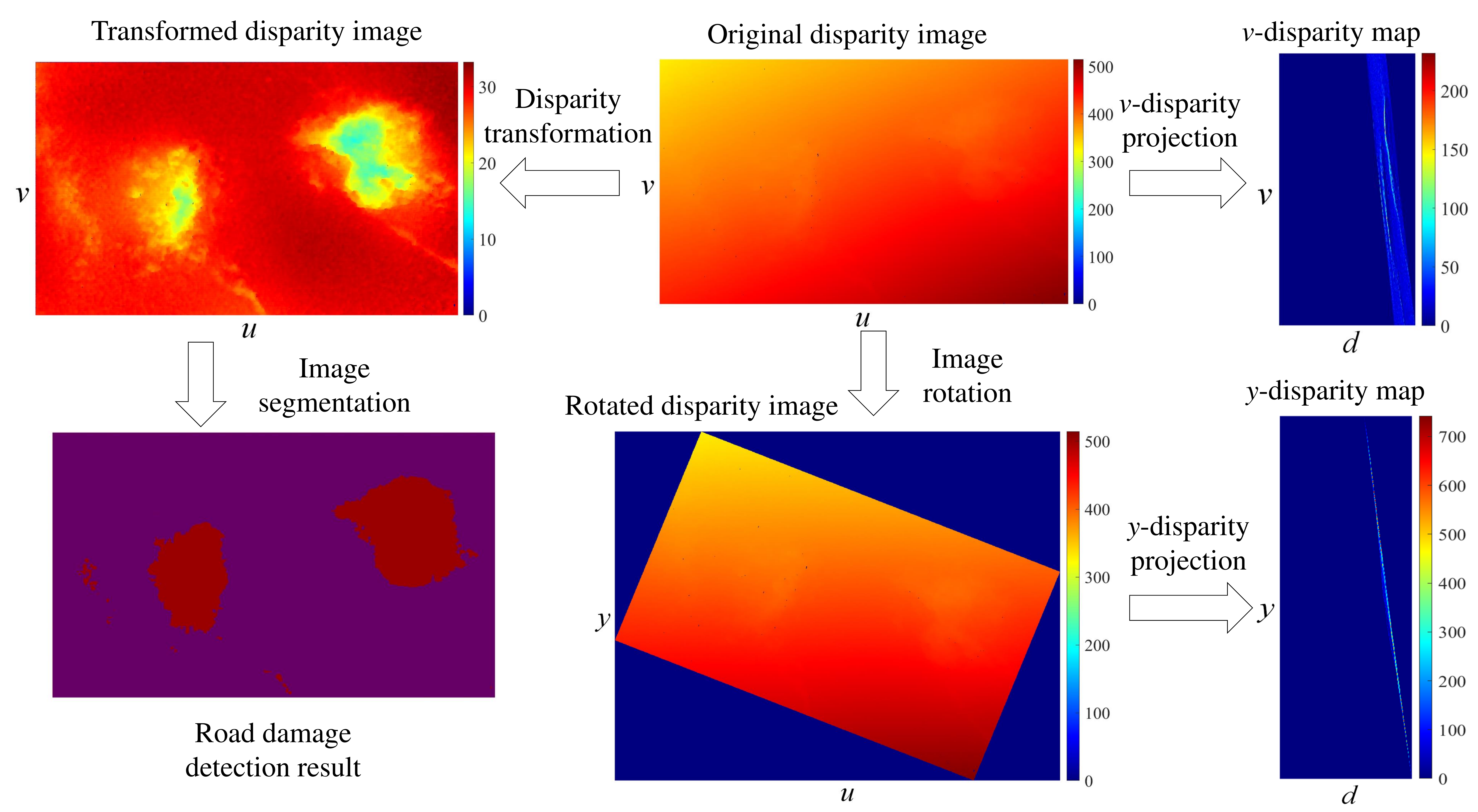}
		\centering
		\caption{Disparity transformation. The stereo rig roll angle $\Phi$ is non-zero, and therefore, the disparity values in the original disparity image change gradually in the horizontal direction. By rotating the original disparity image by $\Phi$, the disparity values on each row become more uniform. Using $\Phi$ and $\mathbf{K}$, the original disparity image can be transformed, where the damaged road areas become highly distinguishable. Finally, the transformed disparity image can be segmented using common image segmentation algorithms for pixel-level road damage detection.}
		\label{fig.disparity_transform}
	\end{center}
\end{figure}

Since the concept of ``v-disparity map'' was presented in \cite{labayrade2002real}, disparity image segmentation has become a common algorithm used for object detection \cite{fan2019road}. In 2018, \cite{fan2018novel} extends the traditional v-disparity representation to a more general form: the projections of the on-road disparity (or inverse depth, because depth is in inverse proportion to disparity) pixels in the v-disparity map can be represented by a non-linear model as follows \cite{fan2021rethinking}:
\begin{equation}
	\tilde{\mathbf{q}}=\mathbf{M}\tilde{\mathbf{p}}={\varkappa}\begin{bmatrix*}[r]
		-\sin\Phi & \cos\Phi & \kappa\\
		0 & {1}/{\varkappa} & 0 \\
		0 & 0 & {1}/{\varkappa}
	\end{bmatrix*}\tilde{\mathbf{p}},
\end{equation}
where $\tilde{\mathbf{p}}=(u;v;1)$ is the homogeneous coordinates of a pixel in the disparity (or inverse depth) image, $\tilde{\mathbf{q}}=(g;v;1)$ is the homogeneous coordinates of its projection in the v-disparity map, $\Phi$ is the stereo rig roll angle, and $\mathbf{K}=(\kappa; \varkappa)$ stores two road disparity projection model coefficients. Additionally,  \cite{fan2018novel} also introduces a novel disparity image processing algorithm, referred to as disparity transformation. This algorithm can not only estimate $\Phi$ and $\mathbf{K}$ but also correct the disparity projections on the v-disparity image. As shown in Fig. \ref{fig.disparity_transform}, the original disparity image can then be transformed to better distinguish between damaged and undamaged road areas, making road damage detection a much easier task.
The disparity transformation is achieved using $\Phi$ and  $\mathbf{K}$, which was estimated using golden section search \cite{pedregal2006introduction} and dynamic programming \cite{fan2018mst, ozgunalp2016multiple}, respectively. Later on, in \cite{fan2019cvprw}, the authors introduced a more efficient solution to $\Phi$, as follows:
\begin{equation}
	\underset{\Phi}{\arg\min}\  \mathbf{g}^\top\mathbf{g}-\mathbf{g}^\top\mathbf{T}(\Phi)\big(\mathbf{T}(\Phi)^\top\mathbf{T}(\Phi)\big)^{-1}\mathbf{T}(\Phi)^\top\mathbf{g},
	\label{eq.phi_eq}
\end{equation}
where $\mathbf{g}$ is a $k$-entry vector of disparity (or inverse depth) values, $\mathbf{1}_k$ is a $k$-entry vector of ones, $\mathbf{u}$ and $\mathbf{v}$ are two $k$-entry vectors storing the horizontal and vertical coordinates of the observed pixels, respectively, and $\mathbf{T}(\Phi)=[\mathbf{1}_k,\cos\Phi \mathbf{v}-\sin\Phi \mathbf{u}]$.  
(\ref{eq.phi_eq}) was solved using gradient descent algorithm \cite{fan2019robust}. In 2019, \cite{fan2019road} proves that (\ref{eq.phi_eq}) has a closed-form solution\footnote{Source code is publicly available at \url{github.com/ruirangerfan/unsupervised_disparity_map_segmentation}} as follows \cite{fan2019road}:
\begin{equation}
	\Phi=
	\arctan\frac{\omega_4\omega_0-\omega_3\omega_1+q\sqrt{\Delta}}{\omega_3\omega_2+\omega_5\omega_1-\omega_5\omega_0-\omega_4\omega_2}
	\ \ \text{s.t.} \ q\in\{-1,1\},
	\label{eq.phi}
\end{equation}
where
\begin{equation}
	\Delta=(\omega_4\omega_0-\omega_3\omega_1)^2+(\omega_3\omega_2-\omega_5\omega_0)^2-(\omega_4\omega_2-\omega_5\omega_1)^2.
	\label{eq.delta}
\end{equation}
The expressions of $\omega_0$-$\omega_5$ are given in \cite{fan2019road}. $\kappa$ and $\varkappa$ can then be obtained using:
\begin{equation}
	\mathbf{x}=	\varkappa\begin{bmatrix}
		\kappa \\
		1
	\end{bmatrix}=\big(\mathbf{T}(\Phi)^\top\mathbf{T}(\Phi)\big)^{-1}\mathbf{T}(\Phi)^\top\mathbf{g}.
\end{equation}
Disparity transformation can therefore be realized using \cite{fan2019road}:
\begin{equation}
	\mathbf{G}'(\mathbf{p})=\mathbf{G}(\mathbf{p})-\varkappa\big(\cos\Phi v - \sin\Phi u \big) - \varkappa\kappa + \Lambda,
\end{equation}
where $\Lambda$ is a constant used to ensure that the values in the transformed disparity (or inverse depth) image $\mathbf{G}'$ are non-negative. As illustrated in Fig. \ref{fig.disparity_transform}, the road damages become highly distinguishable after disparity transformation, and they can be detected with common image segmentation techniques. For instance, \cite{fan2021rethinking} applies superpixel clustering \cite{achanta2012slic} to group the transformed disparities into a set of perceptually meaningful regions, which are then utilized to replace the rigid pixel grid structure. The damaged road areas can then be segmented by selecting the superpixels whose transformed disparity values are lower than those of the healthy road areas. 

\subsubsection{Convolutional neural network-based approaches}

On the other hand, the CNN-based approaches typically address road damage detection in an end-to-end manner, with image classification \cite{hu2019self}, semantic segmentation, or object detection networks.

\paragraph{\textbf{Image classification-based approaches}}

Since 2012, various CNNs have been proposed to solve image classification problems. 
\alexnet \cite{alexnet} is one of the modern CNNs that pushed image classification accuracy significantly. {\vgg} architectures \cite{vgg} improved over \alexnet \cite{alexnet} by increasing the network depth, which enabled them to learn more complicated image features. However, {\vgg} architectures consist of hundreds of millions of parameters, making them very memory-consuming. 
\googlenet \cite{googlenet} (also known as {\inception}-v1) and \inception-v3 \cite{inception} go deeper in parallel paths with different receptive field sizes, so that the {\inception} module can act as a multi-level image feature extractor. Compared to {\vgg} \cite{vgg} architectures, \googlenet \cite{googlenet} and \inception-v3 \cite{inception} have lower computational complexity. However, with the increase of network depth, accuracy gets saturated and then degrades rapidly \cite{resnet}, due to vanishing gradients. To tackle this problem, \cite{resnet} introduces residual neural network ({\resnet}). It is comprised of a stack of building blocks, each of which utilizes identity mapping (skip connection) to add the output from the previous layer to the layer ahead. This helps gradients propagate. In recent years, researchers have turned their focus to light-weight CNNs, which can be embedded in mobile devices to perform real-time image classification. {\mobilenet}-v2 \cite{mobilenetv2}, {\shufflenet}-v2 \cite{shufflenetv2}, and {\mnasnet} \cite{mnasnet} are three of the most popular CNNs of this kind. \mobilenet-v2 \cite{mobilenetv2} mainly introduces the inverted residual and linear bottleneck layer, which allows high accuracy and efficiency in mobile vision applications. 

\begin{figure}[!h]
	\begin{center}
		\centering
		\includegraphics[width=0.98\textwidth]{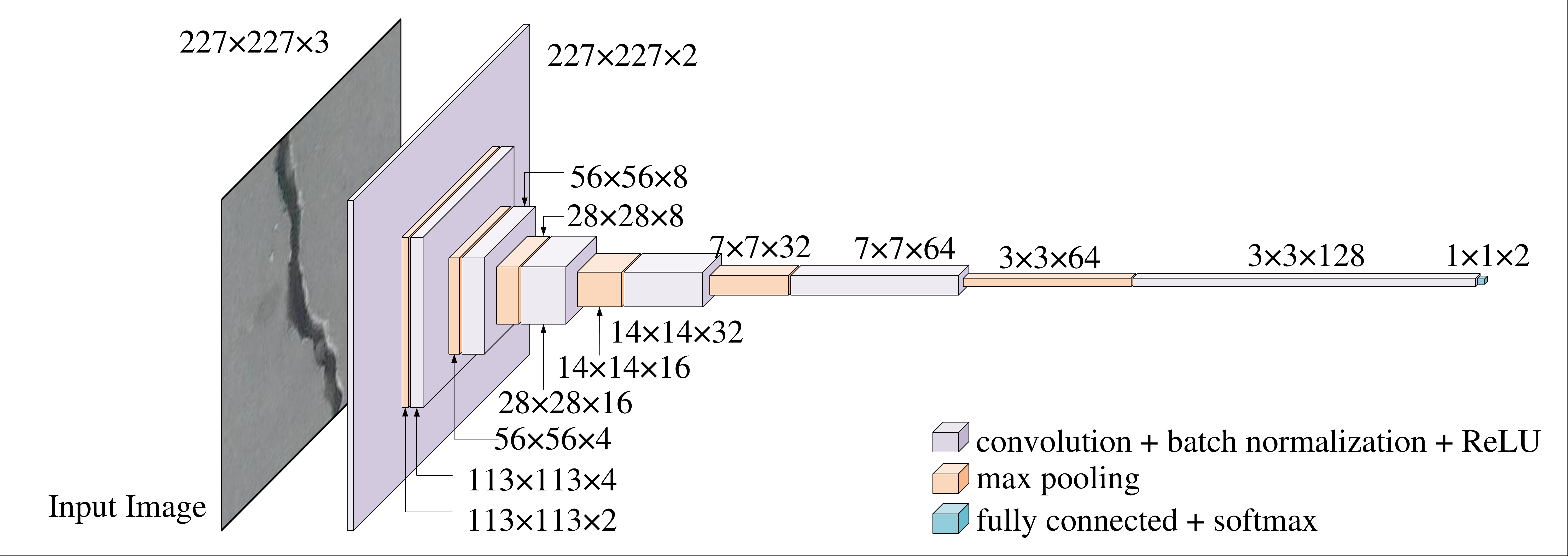}
		\centering
		\caption{The architecture of the road crack detection network presented in \cite{fan2019iv}. }
		\label{fig.cnn_rui_iv}
	\end{center}
\end{figure}

The above-mentioned image classification networks have been widely used to detect, \ie, recognize and localize, road cracks, as the visual features learned by CNNs can replace the traditional hand-crafted features \cite{lecun2015deep}. For example, in 2016, \cite{zhang2016road} proposes a robust road crack detection network. An RGB road image is fed into a CNN consisting of a collection of convolutional layers. The learned visual feature is then connected with a fully connected (FC) layer to produce a scalar indicating the probability that the image contains road cracks. Similarly, in 2019, \cite{fan2019iv} proposes a deep CNN, as illustrated in Fig. \ref{fig.cnn_rui_iv}, which is capable of classifying road images as either positive or negative. The road images are also segmented using an image thresholding method for pixel-level road crack detection. 

In 2019, \cite{hu2019self} proposed a self-supervised monocular road damage detection algorithm, which can not only reconstruct the 3-D road geometry models with multi-view images captured by a single movable camera (based on the hypothesis that the road surface is nearly planar) but also classify RGB road images as either damaged or undamaged with a classification CNN (the images used for training was automatically annotated via a 3-D road point cloud thresholding algorithm). 

\begin{figure}[!h]
	\begin{center}
		\centering
		\includegraphics[width=0.90\textwidth]{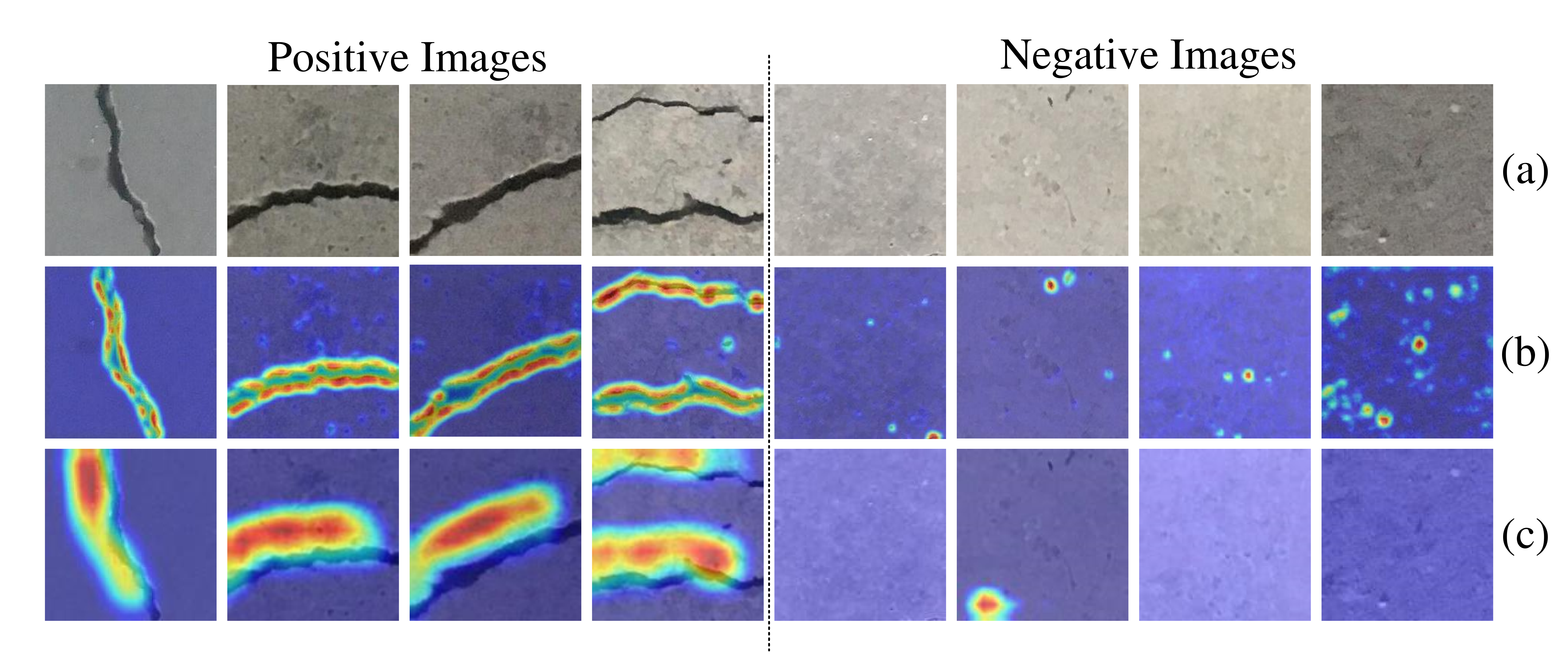}
		\centering
		\caption{Examples of Grad-CAM++ \cite{chattopadhay2018grad} results: (a) road crack images, (b) the class activation maps of ResNet-50 \cite{resnet}, (c) the class activation maps of SENet \cite{hu2018squeeze}. The warmer the color is, the more attention the CNN pays.}
		\label{fig.cam}
	\end{center}
\end{figure}

Recently, \cite{fan2021ist} conducted a comprehensive comparison among 30 SOTA image classification CNNs for road crack detection. The qualitative and quantitative experimental results suggest that road crack detection is not a challenging image classification task. The performances achieved by these deep CNNs are very similar \cite{fan2021ist}. Furthermore, learning road crack detection does not require a large amount of training data. It is demonstrated that 10,000 images are sufficient to train a well-performing CNN \cite{fan2021ist}. However, the pre-trained CNNs typically perform unsatisfactorily on additional test sets. Therefore, unsupervised domain adaptation (UDA) \cite{hoffman2018cycada} capable of mapping two different domains becomes a hot research topic that requires more attention. Finally, explainable AI algorithms, such as Grad-CAM++ \cite{chattopadhay2018grad}, become essential for image classification applications in terms of understanding CNN's decision-making, as shown in Fig. \ref{fig.cam}. 

\paragraph{\textbf{Object detection-based approaches}}

Region-based CNN (R-CNN) \cite{girshick2014rich, girshick2015fast, ren2015faster} series and you only look once (YOLO) \cite{redmon2016you, redmon2017yolo9000, redmon2018yolov3, bochkovskiy2020yolov4} series are two representative groups of modern deep learning-based object detection algorithms. As discussed above, the SOTA image classification CNNs can classify an input image into a specific category. However, we sometimes hope to know the exact location of a particular object in the image, which requires the CNNs to learn bounding boxes around the objects of interest. A naive but straightforward way to achieve this objective is to classify a collection of patches (split from the original image) as positive (the object present in the patch) or negative (the object absent from the patch). However, the problem with this approach is that the objects of interest can have different spatial locations within the image and different aspect ratios. Therefore, it is usually necessary but costly for this approach to select many regions of interest (RoIs). 

\begin{figure}[!h]
	\begin{center}
		\centering
		\includegraphics[width=0.98\textwidth]{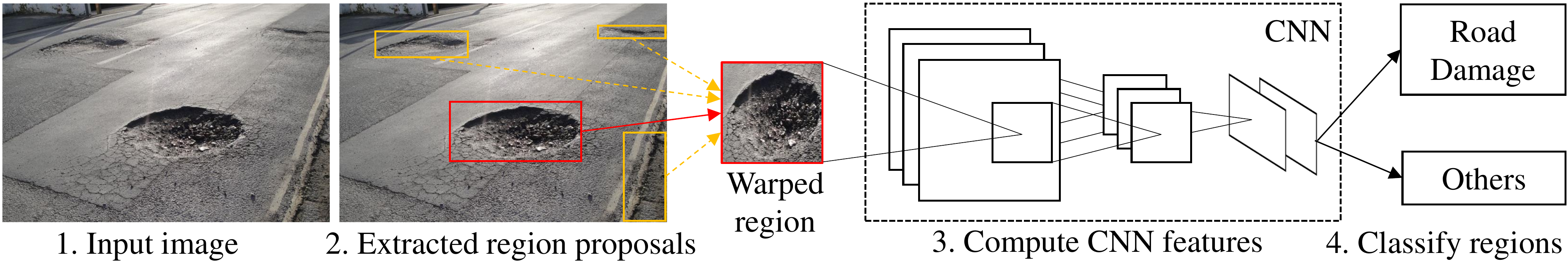}
		\centering
		\caption{R-CNN \cite{girshick2014rich} architecture. }
		\label{fig.rcnn}
	\end{center}
\end{figure}

To solve this dilemma, R-CNN \cite{girshick2014rich} utilizes a selective search algorithm \cite{uijlings2013selective} to extract only 2000 RoIs (referred to as region proposals). Such RoIs are then fed into a classification CNN to extract visual features connected by a support vector machine (SVM) to produce their categories, as illustrated in Fig. \ref{fig.rcnn}. In 2015, Fast R-CNN \cite{girshick2015fast} was introduced as a faster object detection algorithm and is based on R-CNN. Instead of feeding region proposals to the classification CNN, Fast R-CNN \cite{girshick2015fast} directly feeds the original image to the CNN and outputs a convolutional feature map (CFM). The region proposals are identified from the CFM using selective search, which are then reshaped into a fixed size with an RoI pooling layer before being fed into a fully connected layer. A softmax layer is subsequently used to predict the region proposals' classes. However, both R-CNN \cite{girshick2014rich} and Fast R-CNN \cite{girshick2015fast} utilize a selective search algorithm to determine region proposals, which is slow and time-consuming. To overcome this disadvantage, \cite{ren2015faster} proposes Faster R-CNN, which employs a network to learn the region proposals. Faster R-CNN \cite{ren2015faster} is $\sim$250 times faster than R-CNN \cite{girshick2014rich} and $\sim$11 times faster than Fast R-CNN \cite{girshick2015fast}.

On the other hand, the YOLO series are very different from the R-CNN series. YOLOv1 \cite{redmon2016you} formulates object detection as a regression problem. It first splits an image into an $S\times S$ grid, from which $B$ bounding boxes are selected. The network then outputs class probabilities and offset values for these bounding boxes. Since YOLOv1 \cite{redmon2016you} makes a significant number of localization errors and its achieved recall is relatively low, YOLOv2 \cite{redmon2017yolo9000} was proposed to bring various improvements on YOLOv1 \cite{redmon2016you}: (1) it adds batch normalization on all the convolutional layers; (2) it fine-tunes the classification network at the full $448\times448$ resolution; (3) it removes the fully connected layers from YOLOv1 \cite{redmon2016you} and uses anchor boxes to predict bounding boxes. In 2018, \cite{redmon2018yolov3} made several tweaks to further improve \cite{redmon2017yolo9000}. 

As for the application of the aforementioned object detectors in road damage detection, in 2018, \cite{suong2018detection} trained a YOLOv2 \cite{redmon2017yolo9000} object detector to recognize road potholes, while \cite{wang2018road} utilized a Faster R-CNN \cite{ren2015faster} to achieve the same objective. In 2019, \cite{ukhwah2019asphalt} utilized three different YOLOv3 \cite{redmon2018yolov3} architectures to detect road potholes. In 2020, \cite{camilleri2020detecting} leveraged YOLOv3 \cite{redmon2018yolov3} to detect road potholes from RGB images captured by a smartphone mounted in a car. The algorithm was successfully implemented on a Raspberry Pi 2 Model B in TensorFlow. The reported best mean average precision (mAP) was 68.83\% and inference time was 10 ms. Recently, \cite{dhiman2019pothole} compared the performances of YOLOv2 \cite{redmon2017yolo9000} and Mask R-CNN \cite{he2017mask} for road damage detection\footnote{A demo video can be found at \url{vimeo.com/337886918}}. However, these approaches typically employ a well-developed object detection network to detect road damages, without modifications designed specifically for this task. Furthermore, these object detectors can only provide instance-level predictions instead of pixel-level predictions. Therefore, in recent years, semantic image segmentation has become a more desirable technique for road damage detection.

\paragraph{\textbf{Semantic segmentation-based approaches}}

\begin{figure}[!h]
	\begin{center}
		\centering
		\includegraphics[width=0.94\textwidth]{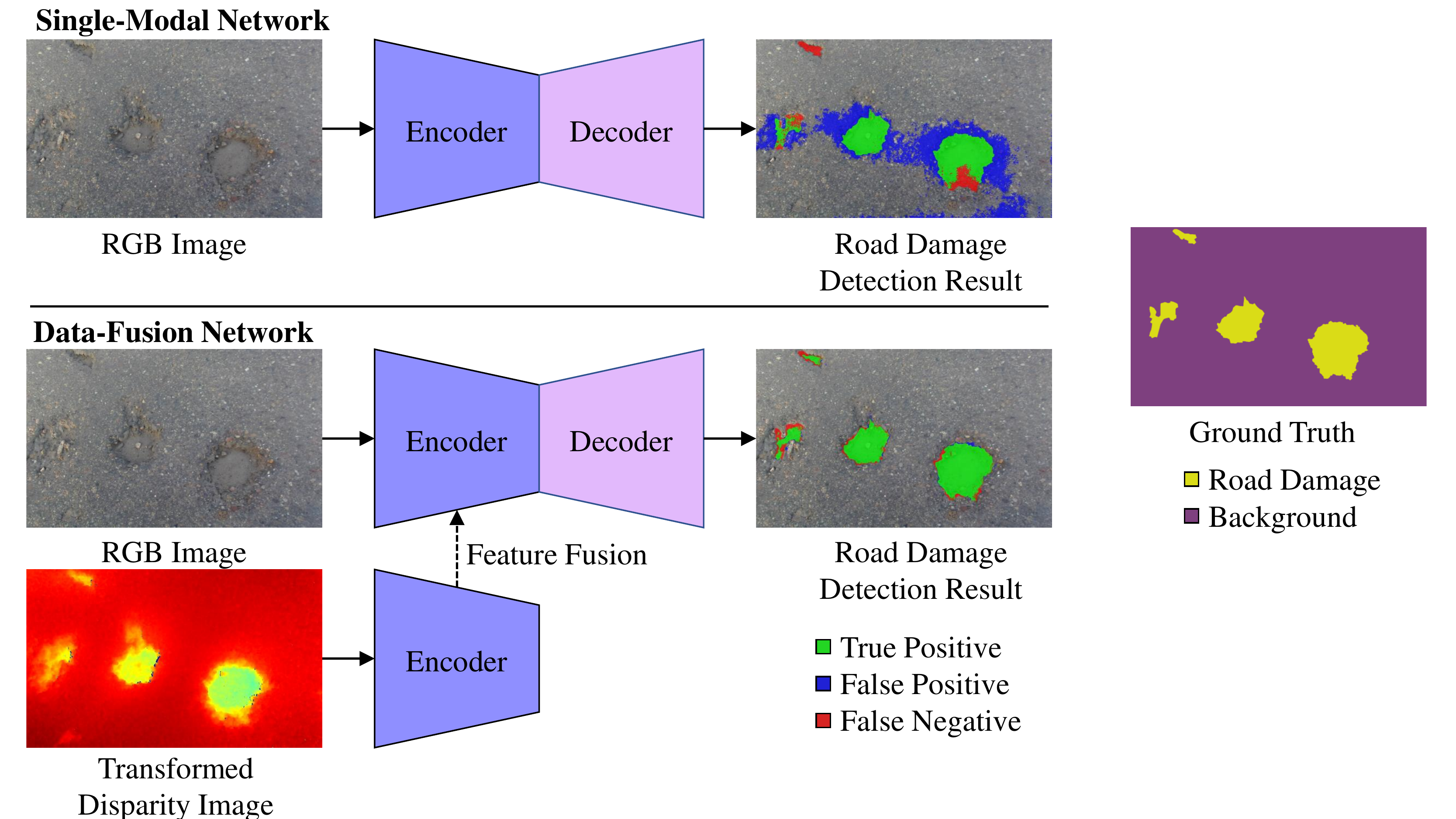}
		\centering
		\caption{Single-modal semantic image segmentation v.s. data-fusion semantic image segmentation for road pothole detection. }
		\label{fig.semantic_seg}
	\end{center}
\end{figure}

Semantic image segmentation aims at assigning each image pixel with a category \cite{fan2020computer}. It has been widely used for road damage detection in recent years. According to the number of input vision data types, the SOTA semantic image segmentation networks can be grouped into two categories: (1) single-modal \cite{fan2021learning} and (2) data-fusion \cite{fan2020sne-roadseg, wang2020sne-roadseg+}, as illustrated in Fig. \ref{fig.semantic_seg}. The former inputs only one type of vision data, such as RGB images or depth images, while the latter typically learns visual features from different types of vision data, such as RGB images and transformed disparity images \cite{wang2021dynamic} or RGB images and surface normal information \cite{fan2020sne-roadseg}. 

Fully convolutional network (FCN) \cite{long2015fully} was the first end-to-end single-modal CNN designed for semantic image segmentation. SegNet \cite{badrinarayanan2017segnet} was the first to use the encoder-decoder architecture, which is widely used in current networks. It consists of an encoder network, a corresponding decoder network, and a final pixel-wise classification layer. U-Net also \cite{ronneberger2015u} employs an encoder-decoder architecture. It adds skip connections between the encoder and decoder to smoothen the gradient flow and restore the object locations. Furthermore, PSPNet \cite{zhao2017pyramid}, DeepLabv3+ \cite{chen2018encoder} and DenseASPP \cite{yang2018denseaspp} leverage a pyramid pooling module to extract context information for better segmentation performance. Additionally, GSCNN \cite{takikawa2019gated} employs a two-branch framework consisting of (1) a shape branch and (2) a regular branch, which can effectively improve the semantic predictions on the boundaries. 

On the other hand, data-fusion networks improve semantic segmentation accuracy by extracting and fusing the features from multi-modalities of visual information \cite{wang2020applying}. For instance, FuseNet \cite{hazirbas2016fusenet} and depth-aware CNN \cite{wang2018depth} adopt the popular encoder-decoder architecture, but employ different operations to fuse the feature maps obtained from the RGB and depth branches. SNE-RoadSeg \cite{fan2020sne-roadseg} and SNE-RoadSeg+ \cite{wang2020sne-roadseg+} extract and fuse the visual features from RGB images and surface normal information (translated from depth/disparity images in an end-to-end manner) for road segmentation. 

Back to the road inspection applications, researchers have already employed both single-modal and data-fusion semantic image segmentation networks to detect road damages/anomalies (the approaches designed to detect road damages can also be utilized to detect road anomalies, as these two types of objects are below and above the road, respectively). 

\begin{figure}[!h]
	\begin{center}
		\centering
		\includegraphics[width=0.92\textwidth]{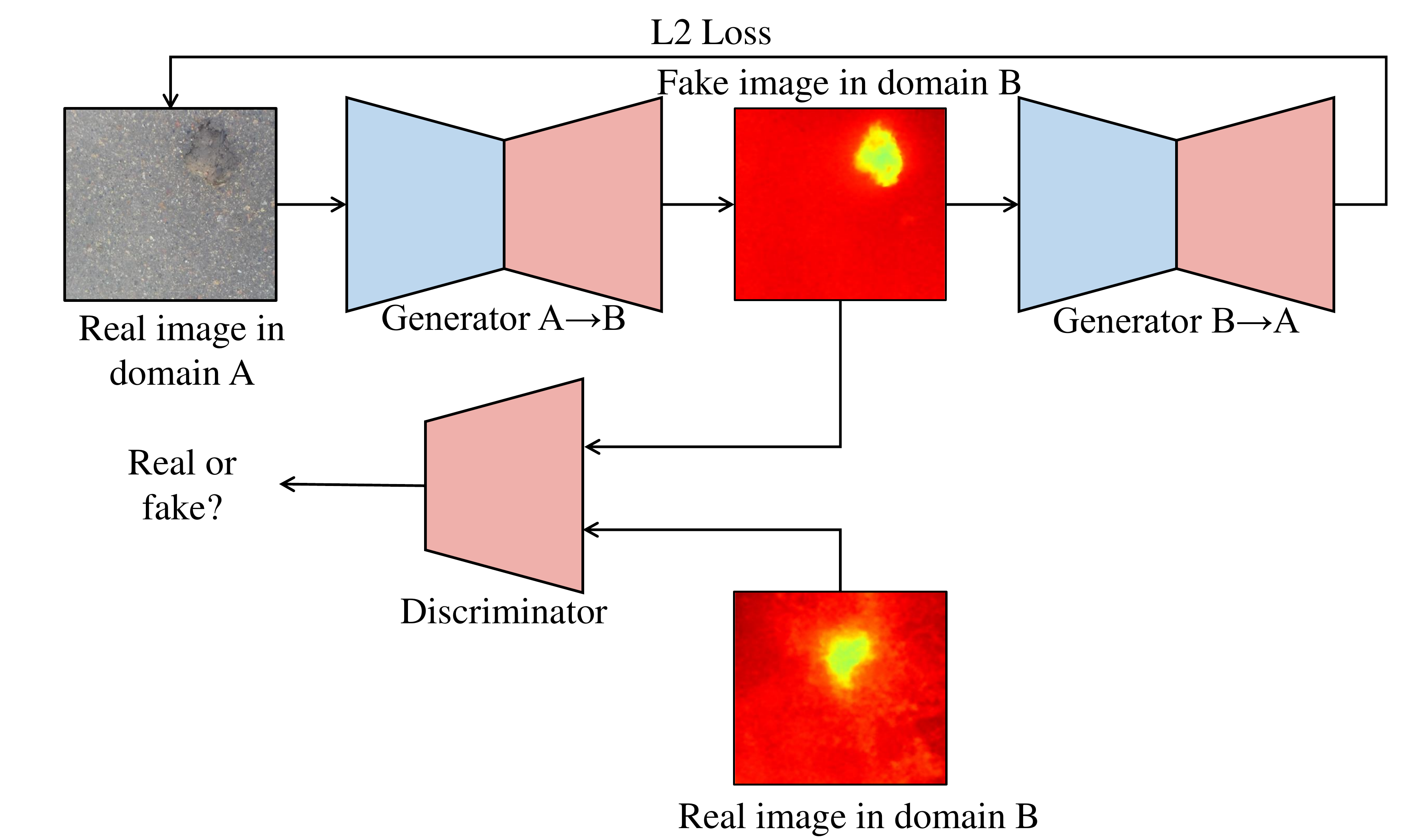}
		\centering
		\caption{CycleGAN \cite{zhu2017unpaired} architecture. }
		\label{fig.gan}
	\end{center}
\end{figure}

In 2020, \cite{fan2020we} proposed a novel attention aggregation (AA) framework\footnote{Source code is publicly available at: \url{github.com/hlwang1124/AAFramework}}, making the most of three different attention modules: (1) channel attention module (CAM), (2) position attention module (PAM), and (3) dual attention module (DAM). Furthermore, \cite{fan2020we} introduces an effective training set augmentation technique based on adversarial domain adaptation (developed based on CycleGAN \cite{zhu2017unpaired}, as illustrated in Fig. \ref{fig.gan}), where the synthetic road RGB images and transformed road disparity (or inverse depth) images are generated to enhance the training of semantic segmentation networks. Extensive experiments conducted with five SOTA single-modal CNNs and three data-fusion CNNs demonstrated the effectiveness of the proposed AA framework and training set augmentation technique. Moreover, a large-scale multi-modal pothole detection dataset (containing RGB images and transformed disparity images), referred to as {Pothole-600}\footnote{\url{sites.google.com/view/pothole-600}}, was published for research purposes. 

\begin{figure}[!h]
	\begin{center}
		\centering
		\includegraphics[width=0.98\textwidth]{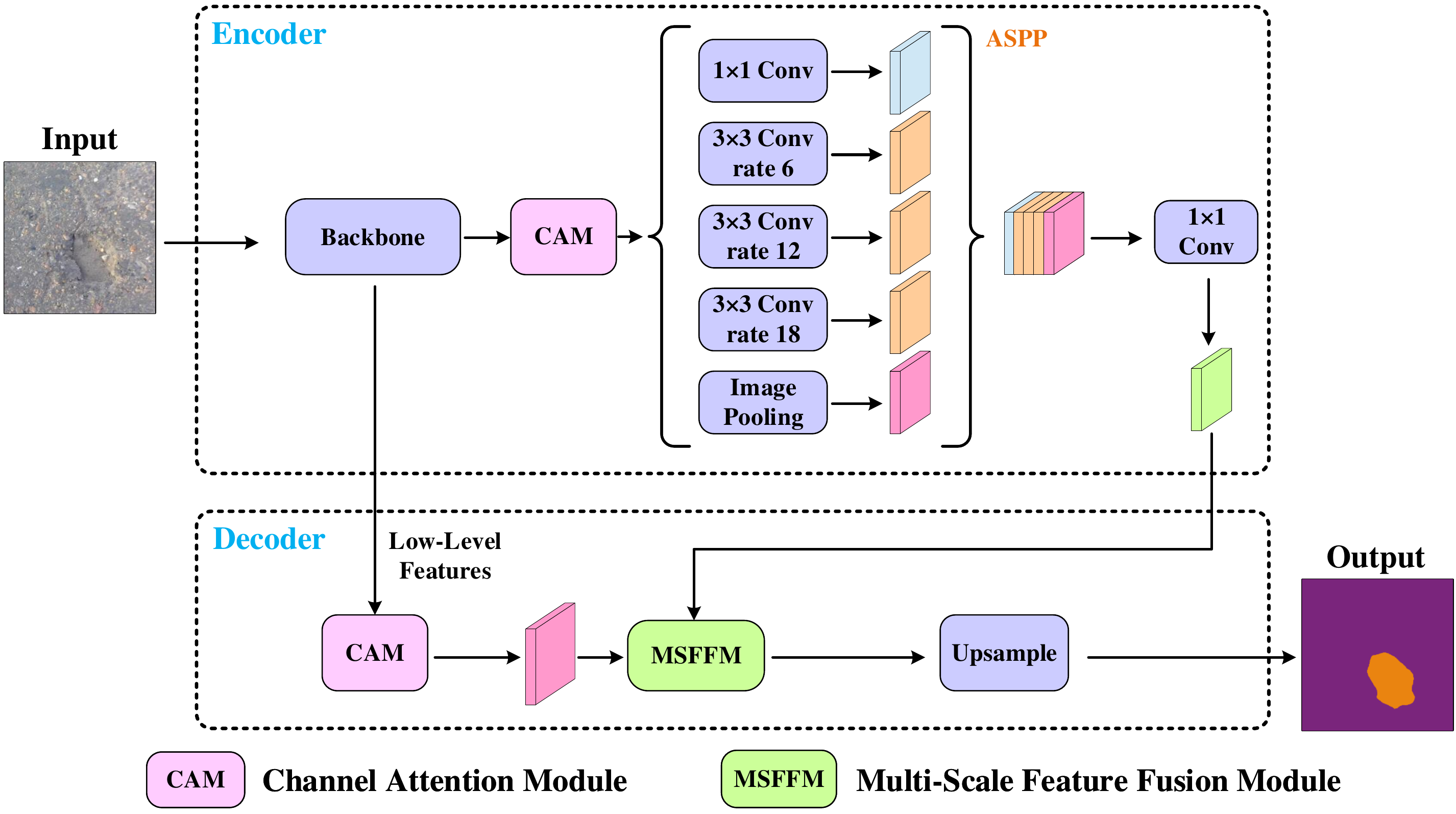}
		\centering
		\caption{The architecture of the road pothole detection network introduced in \cite{fan2021multi}. }
		\label{fig.icas_schema}
	\end{center}
\end{figure}

Similar to \cite{fan2020we}, \cite{fan2021multi} also introduced a road pothole detection approach based on single-modal semantic image segmentation in 2021. Its network architecture, as shown in Fig. \ref{fig.icas_schema}, is developed based on DeepLabv3 \cite{deeplabv3}. It first extracts visual features from an input (RGB or transformed disparity) road image using a backbone CNN. A channel attention module then reweighs the channel features to enhance the consistency of different feature maps. Then, an atrous spatial pyramid pooling (ASPP) module (comprising of atrous convolutions in series, with progressive rates of dilation) is employed to integrate the spatial context information.  This greatly helps to better distinguish between potholes and undamaged road areas. Finally, the feature maps in the adjacent layers are fused using the proposed multi-scale feature fusion module. This further reduces the semantic gap between different feature channel layers. Extensive experimental results conducted on the Pothole-600 \cite{fan2020we} dataset show that the pixel-level mIoU and mFsc are $72.75\%$ and $84.22\%$, respectively. 

Recently, \cite{wang2021dynamic} developed a data-fusion semantic segmentation CNN for road anomaly detection and the authors also conducted a comprehensive comparison of data fusion performance with respect to six different modalities of visual features, including (1) RGB images, (2) disparity images, (3) surface normal images \cite{fan2021three}, (4) elevation images, (5) HHA images (having three channels: (a) disparity, (b) height of the pixels, and (3) angle between the normals and gravity vector based on the estimated ground \cite{hazirbas2016fusenet}), and (6) transformed disparity images \cite{fan2018novel}. The experimental results demonstrated that the transformed disparity image is the most informative visual feature.

\subsection{3-D Road Surface Modeling-Based Approaches}
\label{sec.3d_modeling_approaches}
In 2020, \cite{ravi2020highway} proposed a LiDAR-based road inspection system, where the 3-D road points are classified as damaged and undamaged by comparing their distances to the best-fitting planar 3-D road surface. Unfortunately, \cite{ravi2020highway} lacks the algorithm details and necessary quantitative experimental road damage detection results. 

\begin{figure}[!h]
	\begin{center}
		\centering
		\includegraphics[width=0.50\textwidth]{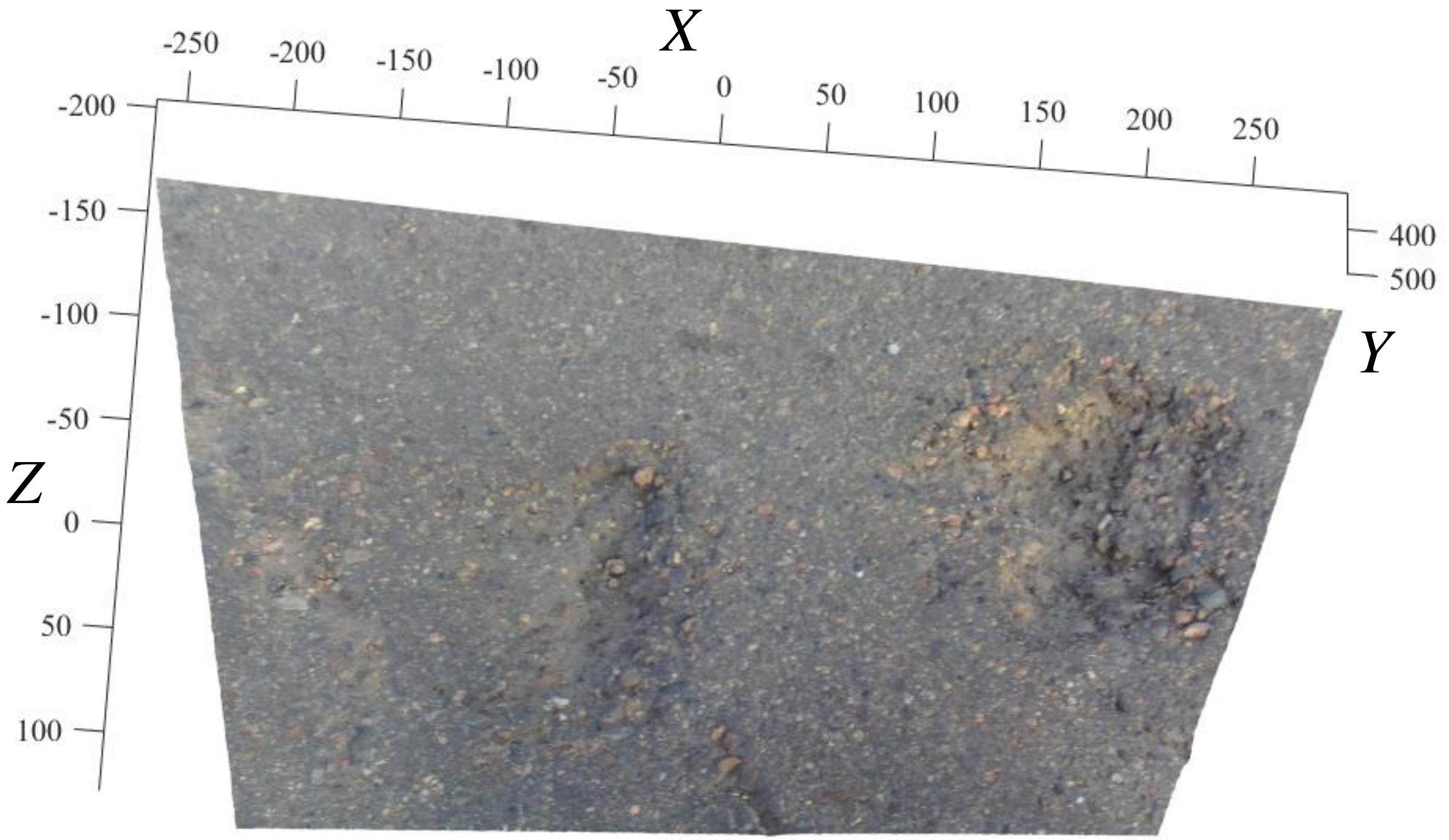}
		\centering
		\caption{An example of 3-D road pothole cloud acquired using the stereo vision technique presented in \cite{fan2018road}. The accuracy of the 3-D road point cloud is 3 mm.}
		\label{fig.3d_point_cloud}
	\end{center}
\end{figure}

In \cite{ozgunalp2016vision, zhang2013advanced, wu2021ist}, the 3-D road point clouds\footnote{3-D road point clouds are publicly available at \url{github.com/ruirangerfan/stereo_pothole_datasets}}, as shown in Fig. \ref{fig.3d_point_cloud},
are formulated as quadratic surfaces, as follows: 
\begin{equation}
	f(X,Z)=a_0+a_1X+a_2Z+a_3X^2+a_4Z^2+a_5XZ,
	\label{eq.surface_modeling}
\end{equation}
where $\mathbf{P}=(X;Y;Z)$ is a 3-D point on the road surface in the camera coordinate system and $\mathbf{a}=(a_0;a_1;a_2;a_3;a_4;a_5)$ stores the quadratic surface coefficients. $\mathbf{a}$ can be estimated by minimizing:
\begin{equation}
	E=\sum_{q=1}^{K}
	\bigg(Y_q-    f(X_q,Z_q)         \bigg)^2.
	\label{eq.E}
\end{equation}
The optimal $\mathbf{a}$ can be obtained when \cite{ozgunalp2016vision}:
\begin{equation}
	\frac{\partial E}{\partial a_0}=\frac{\partial E}{\partial a_1}=\frac{\partial E}{\partial a_2}=\frac{\partial E}{\partial a_3}=\frac{\partial E}{\partial a_4}=\frac{\partial E}{\partial a_5}=0,
\end{equation}
which results in:
\begin{equation}
	\mathbf{M}
	\mathbf{a}=
	\mathbf{q},
\end{equation}
where
\begin{equation}
	\mathbf{M}=	
	\begin{bmatrix}
		K & S_X & S_Z & S_{X^2} & S_{Z^2} & S_{XZ}\\
		S_X & S_{X^2} & S_{XZ} & S_{X^3} & S_{XZ^2} & S_{ZX^2}\\
		S_Z & S_{XZ} & S_{Z^2} & S_{ZX^2} & S_{Z^3} & S_{XZ^2}\\
		S_{X^2} & S_{X^3} & S_{ZX^2} & S_{X^4} & S_{X^2Z^2} & S_{ZX^3}\\
		S_{Z^2} & S_{XZ^2} & S_{Z^3} & S_{X^2Z^2} & S_{Z^4} & S_{XZ^3}\\
		S_{XZ} & S_{X^2Z} & S_{XZ^2} & S_{X^3Z} & S_{XZ^3} & S_{X^2Z^2}
	\end{bmatrix},
	\label{eq.M}
\end{equation}
and
\begin{equation}
	\mathbf{q}=(S_Y;S_{XY};S_{YZ};S_{YX^2};S_{YZ^2};S_{XYZ}).
	\label{eq.q}
\end{equation}
$S$ represents sum operation. For example, $S_{XZ^2}=\sum_1^KX_q{Z_q}^2$ and $S_{XYZ}=\sum_1^KX_q{Y_q}{Z_q}$. 
(\ref{eq.M}) and (\ref{eq.q}) can be rewritten as:
\begin{equation}
	\mathbf{M}=\mathbf{W}^\top\mathbf{W},
\end{equation}
and
\begin{equation}
	\mathbf{q}=\mathbf{W}^\top\mathbf{y},
\end{equation}
where
\begin{equation}
	\mathbf{W}=
	\begin{bmatrix}
		1 & X_1 & Z_1 & {X_1}^2 & {Z_1}^2 & X_1Z_1\\
		\vdots & \vdots & \vdots & \vdots & \vdots & \vdots\\
		1 & X_K & Z_K & {X_K}^2 & {Z_K}^2 & X_KZ_K\\
	\end{bmatrix},
	\label{eq.W}
\end{equation}
and
\begin{equation}
	\mathbf{y}=(Y_1;Y_2;\cdots;Y_n).
\end{equation}
Therefore, $\mathbf{a}$ has the following expression:
\begin{equation}
	\mathbf{a}=\Big( \mathbf{W}^\top \mathbf{W}   \Big)^{-1}\mathbf{W}^\top\mathbf{y}.
	\label{eq.a}
\end{equation}
Plugging (\ref{eq.a}) into (\ref{eq.surface_modeling}) and comparing the difference between the actual and modeled road surfaces, the damaged road areas can be extracted. Nevertheless, 3-D road surface modeling is very sensitive to noise. Hence, \cite{fan2018thesis,ozgunalp2016vision} incorporates the surface normal information into the road surface modeling process to eliminate outliers. Furthermore, \cite{fan2018thesis} also employs random sample consensus (RANSAC) \cite{hast2013optimal} to further improve its robustness. 

\subsection{Hybrid Approaches}
\begin{figure}[!h]
	\begin{center}
		\centering
		\includegraphics[width=0.80\textwidth]{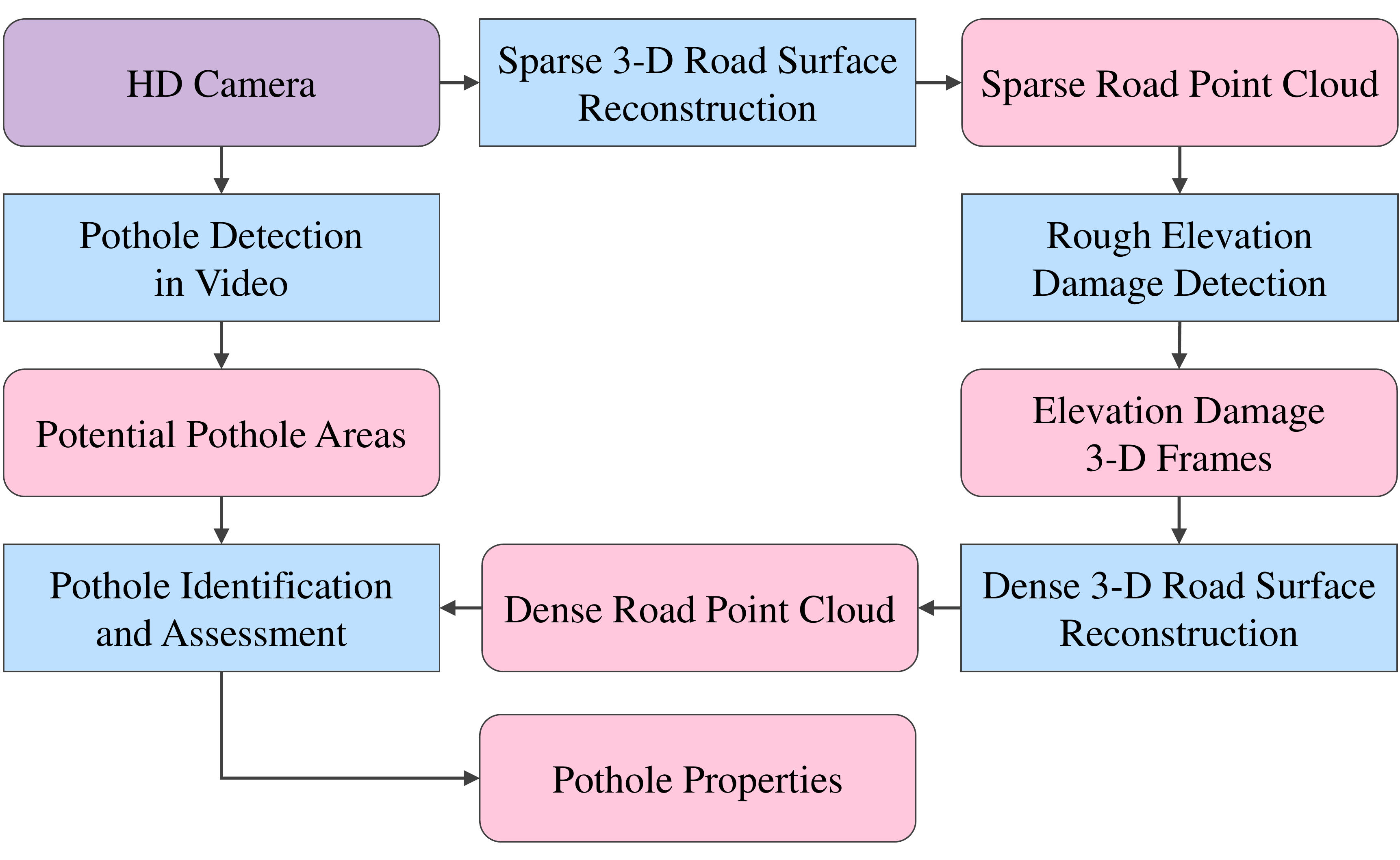}
		\centering
		\caption{The block diagram of the method proposed in\cite{jog2012pothole}. }
		\label{fig.hy1_jog2012pothole}
	\end{center}
\end{figure}

In 2012, \cite{jog2012pothole} introduced a hybrid road pothole detection approach based on 2-D object recognition and 3-D road geometry reconstruction. As shown in Fig. \ref{fig.hy1_jog2012pothole}, the road video frames acquired by a high definition (HD) camera were first utilized to detect (recognize) road potholes. Simultaneously, the same video was also employed for sparse 3-D road geometry reconstruction. By analyzing such multi-modal road inspection results, the potholes were accurately detected. Such a hybrid method dramatically reduces the incorrectly detected road potholes.
\begin{figure}[!h]
	\begin{center}
		\centering
		\includegraphics[width=0.86\textwidth]{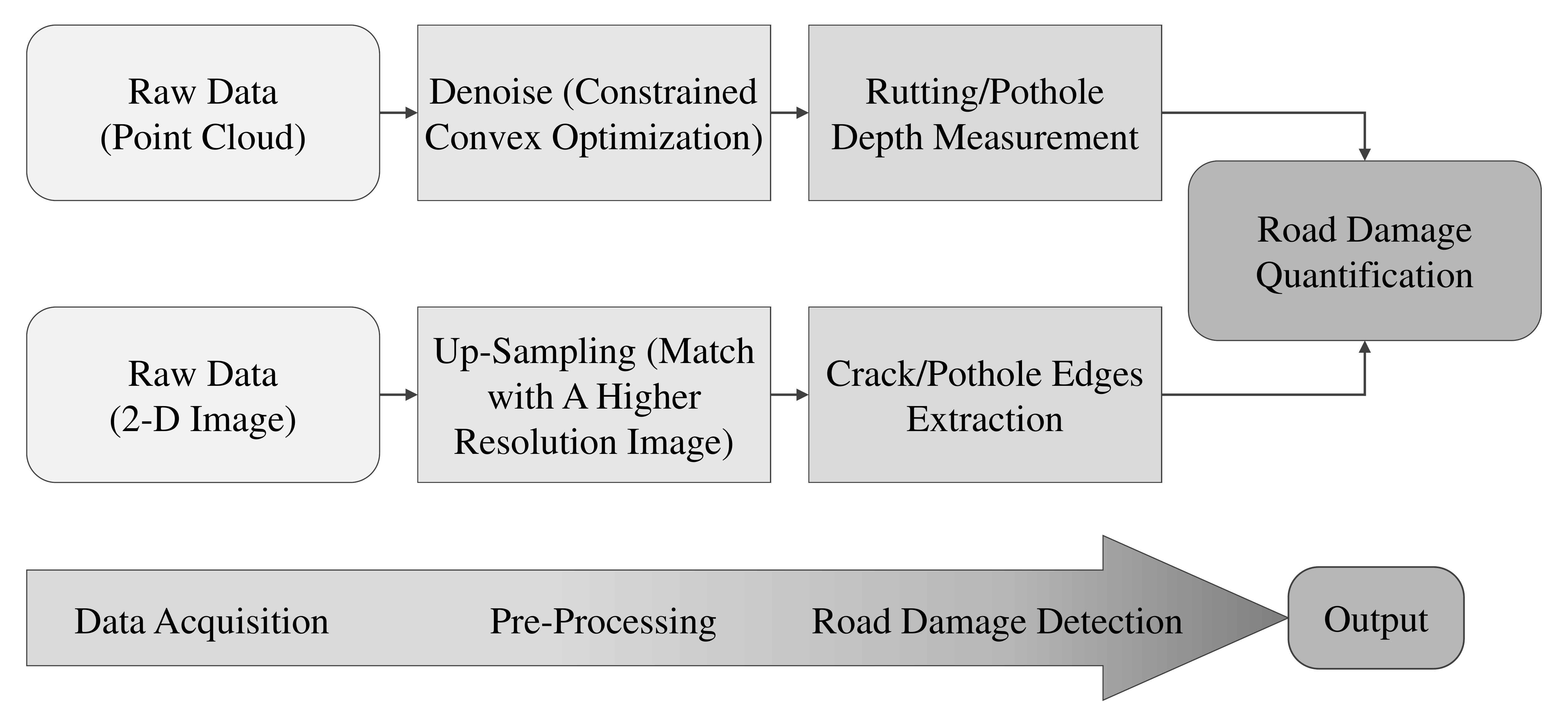}
		\centering
		\caption{The block diagram of the road damage detection method proposed in \cite{yuan2014automatic}. }
		\label{fig.hy2_yuan2014automatic}
	\end{center}
\end{figure}

In 2014, \cite{yuan2014automatic} also introduced a hybrid road damage detection approach, which can effectively recognize and measure road damages. As illustrated in Fig. \ref{fig.hy2_yuan2014automatic}, this algorithm requires two modalities of road data: (1) 2-D gray-scale/RGB road images (2) 3-D road point clouds. The road images are up-sampled by matching them with higher resolution images. The road pothole/crack edges are then extracted from the up-sampled images. On the other hand, the road point clouds are denoised through constrained convex optimization to measure rutting/pothole depth. Finally, the extracted pothole/crack edges and measured rutting/pothole depth are combined for road damage quantification. Unfortunately, these two modalities of road data were processed independently, and the data fusion process was not discussed in this work. 

In 2017, \cite{kang2017pothole} proposed an automated road pothole detection system based on the analysis of 2-D LiDAR data and RGB road images. This hybrid system has the advantage of not being affected by electromagnetic waves and poor road conditions. The 2-D LiDAR data provide road profile information, while the RGB road images provide road texture information. To obtain a highly accurate and large road area, this system uses two LiDARs. Such a hybrid system can combine the advantages of different sources of vision data to improve the overall road pothole detection accuracy.
\begin{figure}[!h]
	\begin{center}
		\centering
		\includegraphics[width=0.86\textwidth]{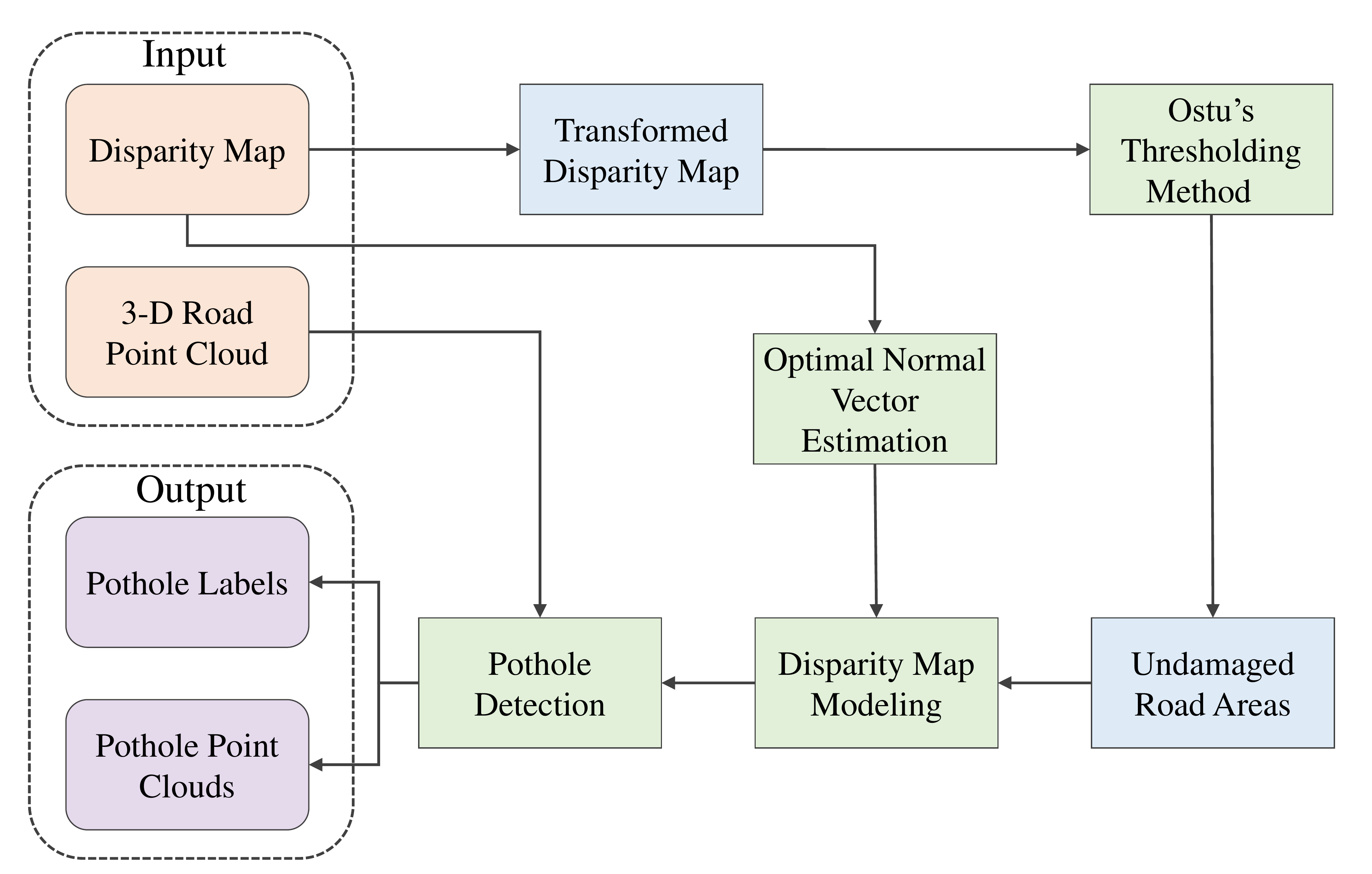}
		\centering
		\caption{The road pothole detection workflow proposed in \cite{fan2019pothole}. }
		\label{fig.hy4_fan2019pothole}
	\end{center}
\end{figure}

In 2019, \cite{fan2019pothole} also introduced a hybrid framework for road pothole detection. The methodology is illustrated in Fig. \ref{fig.hy4_fan2019pothole}. It first applies disparity transformation, discussed in Sec. \ref{sec.traditional_approaches}, and Otsu's thresholding \cite{otsu1979threshold} methods to extract  potential undamaged road areas. In the next step, a quadratic surface was fitted to the original disparity image, where the RANSAC algorithm was employed to enhance the surface fitting robustness. Potholes can be successfully detected by comparing the actual and fitted disparity images. This hybrid framework uses both 2-D image processing and 3-D road surface modeling algorithms,  greatly improving the road pothole detection performance.

\section{Parallel Computing Architecture}
\label{sec.parallel_computing_architecture}
Graphics processing unit (GPU) was initially developed to accelerate graphics processing. It has now evolved as a specialized processing core, significantly speeding up computational processes. GPUs have unique parallel computing architectures and they can be used for a wide range of massively distributed computational processes, such as graphics rendering, supercomputing, weather forecasting, and autonomous driving. 

The main advantage of GPUs is its massively parallel architecture. There are several different ways to classify parallel computers. One of the most widely used taxonomies, in use since 1966, is referred to as Flynn's taxonomy \cite{Flynn1966}. Flynn classifies parallel architectures based on the two independent dimensions of instruction and data. Each dimension can only be either single or multiple. There are four parallel computing architectures: (1) single instruction, single data (SISD), (2) single instruction, multiple data (SIMD), (3) multiple instructions, single data (MISD), and (4) multiple instructions, multiple data (MIMD).

\begin{figure}[!h]
	\begin{center}
		\centering
		\includegraphics[width=0.95\textwidth]{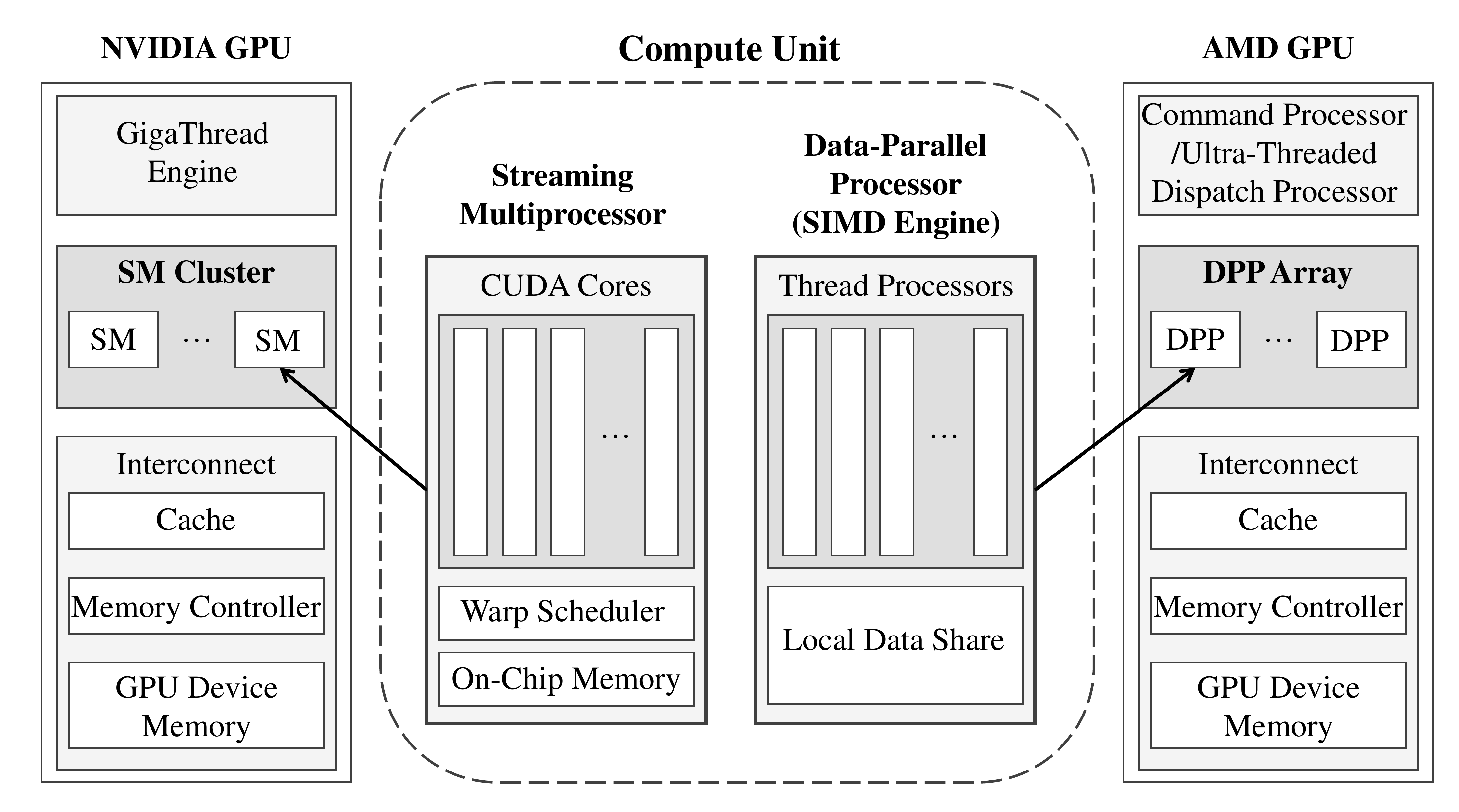}
		\centering
		\caption{NVIDIA and AMD GPU architectures presented in \cite{Lee2013}. }
		\label{fig.gpu_architecture}
	\end{center}
\end{figure}

As illustrated in Fig. \ref{fig.gpu_architecture}, there are currently two main popular GPU architectures \cite{Lee2013}: (1) NVIDIA's Tesla/Fermi GPU architecture and (2) AMD's Evergreen/Northern Island series GPU architecture. The major difference between these two types of GPU architectures is the processor core: the computing unit of NVIDIA GPUs is a streaming multiprocessor (SM) cluster, while the computing unit of AMD GPUs is a data-parallel processor (DPP) array. 

For NVIDIA GPUs, different types of GPUs may have different SM numbers. Each SM has 8 to 48 CUDA cores composed of three parts: an integer/floating-point unit, a warp scheduler for instruction scheduling, and on-chip memory. For AMD GPU, its computing unit DPP has 16 thread processors, containing the execution pipeline of the very long instruction word (VLIW) architecture and the local data sharing part. Furthermore, NVIDIA uses a GigaThread engine while AMD uses a command processor/ultra-threaded dispatch processor to control compute units. In addition, the GPU also includes the interconnection part between the computing unit and the memory subsystem. They are all composed of cache, memory controller, and GPU device memory.

Parallelism is the basis of high-performance computing. With the continuous improvement in computing power and parallelism programmability, GPUs have been employed in an increasing number of deep learning applications, such as road condition assessment with semantic segmentation networks, to accelerate model training and inference.

\section{Summary}
\label{sec.summary}
This chapter gave readers an overall picture of the SOTA computer-aided road inspection approaches. Five common types of road damages, including crack, spalling, pothole, rutting, and shoving, were discussed; The 2-D/3-D road imaging techniques, especially laser scanning, IR sensing, and multi-view geometry from digital images, were discussed. Finally, the existing machine vision and intelligence approaches, including 2-D image analysis/understanding algorithms, 3-D road surface modeling algorithms, and the hybrid methods developed to detect road damages were reviewed.

\bibliographystyle{IEEEtran}

\end{document}